\documentclass[letterpaper, 10 pt, conference]{ieeeconf} 

\IEEEoverridecommandlockouts

\usepackage{graphicx} 
\usepackage[utf8]{inputenc}
\usepackage{todonotes} 
\usepackage{amsmath,amssymb,amsfonts,booktabs,cite} 
\usepackage{siunitx,textcomp} 
\usepackage{algorithmic,algorithm} 
\usepackage[hidelinks]{hyperref} 
\usepackage{multicol}
\usepackage{epstopdf} 
\usepackage{subcaption}
\let\labelindent\relax
\usepackage[inline]{enumitem} 
\usepackage[nolist]{acronym} 
\usepackage{colortbl} 	    
\usepackage{overpic}
\usepackage{tikz}
\usetikzlibrary{patterns}
\usepackage{pgfplots}
\pgfplotsset{every axis/.append style={font=\small}}
\pgfplotsset{ylabel near ticks}
\pgfplotsset{xlabel near ticks}
\usepackage{standalone} 
\usepackage{adjustbox}
\pgfplotsset{compat=1.13}
\pgfplotsset{grid style={dashed,gray}}
\pgfplotsset{minor grid style={dashed,red}}
\pgfplotsset{major grid style={dotted,green!50!black}}
\newacro{bn}[BN]{Batch Normalization}
\newacro{relu}[ReLU]{Rectified Linear Unit}
\newacro{adam}[Adam]{Adaptive Moment Estimation}
\newacro{ai}[AI]{Artificial Intelligence}
\newacro{dl}[DL]{Deep Learning}
\newacro{dnn}[DNN]{Deep Neural Network}
\newacro{cnn}[CNN]{Convolutional Neural Network}
\newacro{bnn}[BNN]{Bayesian Neural Network}
\newacro{mc}[MC]{Monte-Carlo}
\newacro{dof}[DoF]{Degrees-of-Freedom}
\newacro{ods}[USN]{Uncertain Shape Network}
\newacro{od}[SN]{Shape Network}
\newacro{gp}[GP]{Gaussian Processes}
\newacro{mcmc}[MCMC]{Markov Chain Mote Carlo}
\newacro{psdf}[p-SDF]{probabilistic Signed Distance Function}
\newacro{gpis}[GPISs]{Gaussian Process Implicit Surfaces}
\newacro{va}[V]{Varley}
\newacro{rl}[RL]{Reinforcemnt Learning}
\newacro{fcgqcnn}[FC-GQ-CNN]{Fully Convolutional Grasp Quality CNN}
\newacro{gqcnn}[GQCNN]{Grasp Quality CNN}
\newacro{otg}[STG]{Simulated Top-Grasps}
\newacro{oag}[SAG]{Simulated All-Grasps}
\newacro{gp}[GP]{Gaussian Process}
\newacro{dof}[dof]{degrees-of-freedom}
\newacroplural{dof}[dof]{degrees-of-freedom}
\newacro{sota}[sota]{state-of-the-art}

\newcommand{\figref}[1]{\hyperref[#1]{Fig.~\ref*{#1}}}
\newcommand{\tabref}[1]{\hyperref[#1]{Table~\ref*{#1}}}
\newcommand{\secref}[1]{\hyperref[#1]{Section~\ref*{#1}}}
\newcommand{\algoref}[1]{\hyperref[#1]{Algorithm~\ref*{#1}}}

\newcommand{\ra}[1]{\renewcommand{\arraystretch}{#1}}
\newcommand{\tbs}[1]{\renewcommand{\tabcolsep}{#1pt}}
\newlength\myindent
\newcommand{\matr}[1]{\mathbf{#1}}
\DeclareMathOperator*{\argmax}{arg\,max}

\newcommand{\E}[1]{\mathbb{E}\left[#1\right]}

\def\bestcolor{(best viewed in color)}
\def\panda{\textit{Franka Emika Panda}}

\def\sota{state-of-the-art}

\def\ie{\textit{i.e.,}}
\def\eg{\textit{e.g.,}}
\def\etal{\textit{et al.}}

\def\pc{point-cloud}

\def\nat{\multicolumn{1}{c}{--}}
\def\mujoco{MuJoCo}
\def\dd{data-driven grasp planning}
\def\etedd{end-to-end data-driven grasp planning}
\def\etefdd{end-to-end data-driven 4 \acp{dof} grasp planning}
\def\depthimage{depth image}
\definecolor{significant}{RGB}{217,217,217}
\definecolor{verysignificant}{RGB}{255,255,255}

\title{\LARGE \bf%
Beyond Top-Grasps Through Scene Completion}

\author{Jens~Lundell, Francesco~Verdoja 
and Ville~Kyrki%
\thanks{This work was supported by the Strategic Research Council at Academy of 
Finland, decision 314180.}
\thanks{J.~Lundell, F.~Verdoja and V.~Kyrki are with School of Electrical 
Engineering, Aalto University, Finland. 
\texttt{\{firstname.lastname\}{@}aalto.fi}}}

\begin{document}

\maketitle
\thispagestyle{empty}
\pagestyle{empty}


\begin{abstract}
Current end-to-end grasp planning methods propose grasps in the order of seconds that attain high grasp success rates on a diverse set of objects, but often by constraining the workspace to top-grasps. In this work, we present a method that allows end-to-end top-grasp planning methods to generate full six-degree-of-freedom grasps using a single RGB-D view as input. This is achieved by estimating the complete shape of the object to be grasped, then simulating different viewpoints of the object, passing the simulated viewpoints to an end-to-end grasp generation method, and finally executing the overall best grasp. The method was experimentally validated on a \panda{} by comparing 429 grasps generated by the \sota{} Fully Convolutional Grasp Quality CNN, both on simulated and real camera images. The results show statistically significant improvements in terms of grasp success rate when using simulated images over real camera images, especially when the real camera viewpoint is angled. Code and video are available  at \href{https://irobotics.aalto.fi/beyond-top-grasps-through-scene-completion/}{https://irobotics.aalto.fi/beyond-top-grasps-through-scene-completion/}.
\end{abstract}

\acresetall
\section{Introduction}
\label{sec:intro}

Robotic grasping has undergone a paradigm shift from analytical methods toward data-driven ones. Deep learning is the major driving force behind the shift and has given rise to a diverse set of methods \cite{mahler2017dex,satish2019policy,aktas2019deep,varley2015generating,johns_deep_2016,choi_learning_2018,bousmalis2018using,tobin2018domain}. These methods typically reach high grasp success rates (often above 90\%) on a wide variety of objects while keeping the total computation time in the order of seconds, surpassing analytical methods by a large margin. However, to reach such a performance the grasp planning problem is usually constrained to the generation of top-grasps with four \acp{dof}: one orientation and three translations. Top-grasps are good if the camera perceiving the environment is perpendicular to the plane supporting the target. However, as shown in this work, once the camera views a scene from an angle the performance drops. In such situations the grasping methods need to propose grasps in full six \acp{dof} space to allow a robot to approach objects from any possible direction.

\begin{figure}
	\centering
	\begin{subfigure}{0.27\textwidth}
		\includegraphics[width=\linewidth]{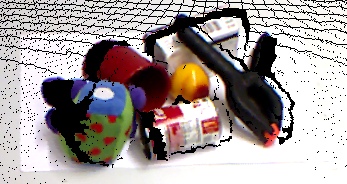}
		\caption{\label{fig:pc}}
	\end{subfigure}\hfil
	\begin{subfigure}{0.21\textwidth}
	\includegraphics[width=\linewidth]{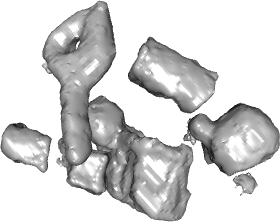}
		\caption{\label{fig:scene_comp}}
\end{subfigure}
\medskip
	\begin{subfigure}{0.485\textwidth}
		\includegraphics[width=\linewidth]{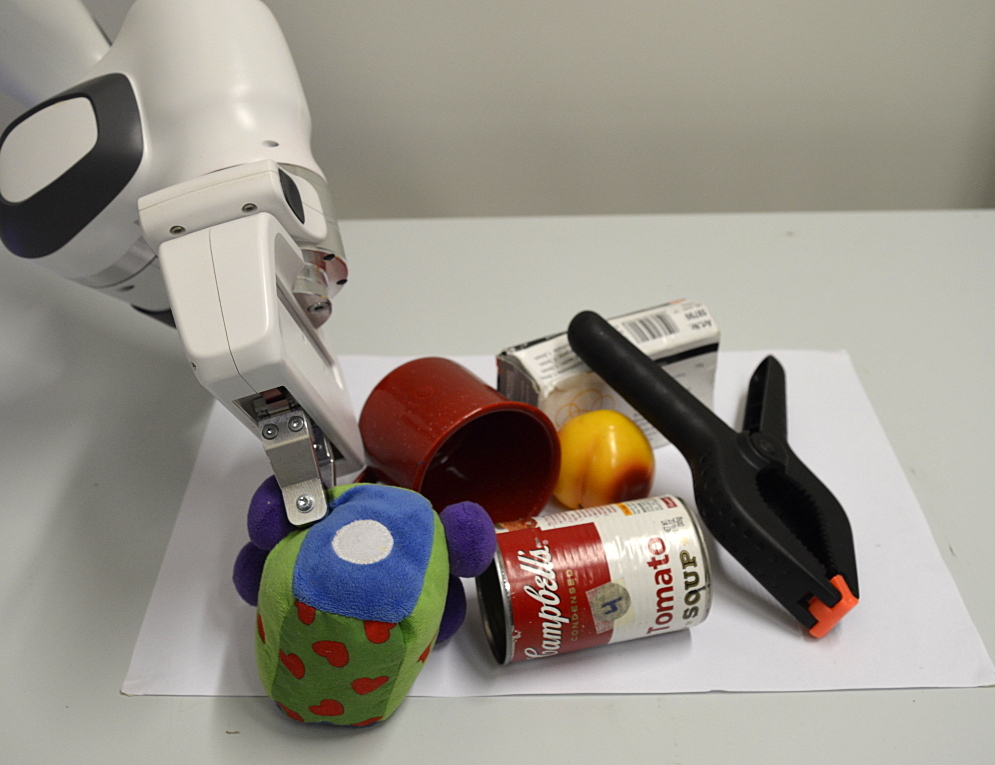}
		\caption{\label{fig:real_grasping}}
	\end{subfigure}
	\caption{(a) shows a \pc{} of a cluttered scene acquired with a real camera while (b) shows a simulated top-view of the same scene but shape completed. (c) shows that with our method, successful grasps can be generated from approach directions different from the camera viewpoint.\label{fig:cluttered_scene}}
	\vspace{-1em}
\end{figure}

One viable option to achieve full 6 \acp{dof} grasping with current \sota{} grasping methods is to mount a camera on the robot itself and have it scan the scene from multiple viewpoints. However, not only is such an approach slow as the robot needs to first plan where to move and then physically move there but also the robot might self-occlude the view of the camera, rendering the method useless. A novel alternative, which is studied in this work, is to simulate different viewpoints of the object to be grasped and feed these to the methods proposing 4 \acp{dof} grasps. As shown in \figref{fig:cluttered_scene}, such a solution enables a robot to grasp objects from directions different from the one of the real camera.

TO this end, we present a grasping pipeline that uses the state-of-the-art \ac{fcgqcnn} \cite{satish2019policy} to propose grasps.
The pipeline first segments a \pc{} of the scene into objects. Then the shape of each object is estimated and placed in a physics simulator.
In the context of this work, we refer to this as \textit{scene completion}.
Inside the physics simulator, a set of \depthimage{}s are sampled from different viewpoints and fed to the grasp proposal method generating a set of grasp candidates.
The grasp candidate with the highest score is then executed on the robot. 

The proposed grasping pipeline is experimentally validated on a \panda{} by benchmarking it against grasps proposed on real \depthimage{}s on single object grasping and grasping in clutter. The results of 429 grasps on single object grasping show statistically significant improvement in terms of higher grasp success rate when planning is performed on simulated \depthimage{}s compared to planning solely on real camera images. Similar results were also evident when grasping in cluttered scenes.

The main contributions of this paper are: 
\begin{enumerate*}[label=(\roman*)]
	\item a novel grasp planning pipeline that enables existing 4 \acp{dof} end-to-end methods to propose full 6 \acp{dof} grasps, 
	\item a method to densely sample simulated \depthimage{}s, and
	\item an empirical evaluation of the proposed method against \sota{} on real hardware, presenting a statistically significant improvement in terms of higher grasp success rate using the proposed method.
\end{enumerate*}

\section{Related Works}
\label{sec:related}

To date, many grasping methods rely fully or in part on deep learning. Some methods only use deep learning to extract additional information about objects with \eg{} \textit{shape completion} \cite{varley_shape_2017,lundell2019robust} or tactile information \cite{watkins-valls_multi-modal_2018} and then use analytical methods to plan the actual grasp \cite{sahbani2012overview}, while others employ \textit{\dd{}} in an end-to-end fashion to generate grasps directly from images \cite{mahler2017dex,satish2019policy,aktas2019deep,varley2015generating,johns_deep_2016,choi_learning_2018,bousmalis2018using,tobin2018domain}. We will review both shape completion and \etedd{} as both are vital parts of our grasping pipeline.

\subsection{Deep Shape Completion}

In the context of shape completion from incomplete \pc{}s, most recent 
improvements come from the adoption of deep learning. For instance, different works have explored tailored network structures 
\cite{varley_shape_2017,han_high-resolution_2017,yang20173d}, semantic object 
classification to aid the reconstruction \cite{dai2017complete}, the 
integration of other sensing modalities such as tactile information 
\cite{watkins-valls_multi-modal_2018}, or the exploitation of the network uncertainty 
\cite{lundell2019robust}.

In the context of robotics grasping, \cite{varley_shape_2017}, 
\cite{watkins-valls_multi-modal_2018} and \cite{lundell2019robust} are the most interesting 
as they not only focus on shape reconstruction quality but also on grasping 
accuracy. In this work, we make use of our previous shape completion network \cite{lundell2019robust} to complete objects but instead of planning grasps with analytical methods---which is computationally expensive---we turn to \dd{}.

\subsection{End-To-End Data-Driven Grasp Planning}

The general interest in \etedd{} came after the pioneering work by Saxena \etal \cite{saxena_robotic_2008} where they trained a logistic regression model to directly predict good grasping points from a monocular image. To train the logistic regressor they used a large  amount of synthetically labeled images of objects and the corresponding grasping location. 

The use of synthetic data to train the sensor-to-grasp map was later used in a wide variety of similar methods \cite{mahler2017dex,satish2019policy,aktas2019deep,varley2015generating,johns_deep_2016,choi_learning_2018,bousmalis2018using,tobin2018domain}. For instance, Mahler \etal{} \cite{mahler2017dex} used a data-set containing millions of synthetic antipodal top-grasps on a wide variety of objects to train a \ac{gqcnn} that generates a grasp from a \depthimage{} in the order of seconds. The \ac{gqcnn} was later improved in \cite{satish2019policy} through the use of on-policy data and a fully convolutional network structure called \ac{fcgqcnn}. The \sota{} \ac{fcgqcnn} was faster than \ac{gqcnn} while sampling about 5000x more grasps and was thus used to generate grasps in this work.

Another line of research in \etedd{} is \ac{rl} \cite{quillen_deep_2018,pinto_supersizing_2015,levine_learning_2016,james2019sim} where the goal is to learn the sensor-to-grasp map directly through trial and error on the robot. Models learned with \ac{rl} can attain a high grasp success rate without any hand-labeled data-sets, but the extensive interaction time needed to learn the model, which can be months on physical robots \cite{levine_learning_2016}, is a bottleneck. Although some work have reduced the real-world interaction by using simulation \cite{james2019sim}, the learned models still needs fine-tuning on physical hardware to reach similar grasp success rate as methods that uses supervision \cite{mahler2017dex,satish2019policy}.

A major limitation in most \etedd{} works is that the planned grasp is only from the viewpoint of the camera, which effectively constrains the grasps to a subset of the complete 6 \acp{dof} workspace (typically 4 \acp{dof} grasps are considered). The work presented here lifts this limitation by shape completing the real objects, placing them into a physics simulator, and from there sampling different viewpoints of the object. Our method hence enables standard \etedd{} methods that suggest grasps from only one camera viewpoint to generate full 6 \acp{dof} grasps from other directions such as the back of the object.

\section{Problem formulation}
\label{sec:problem}

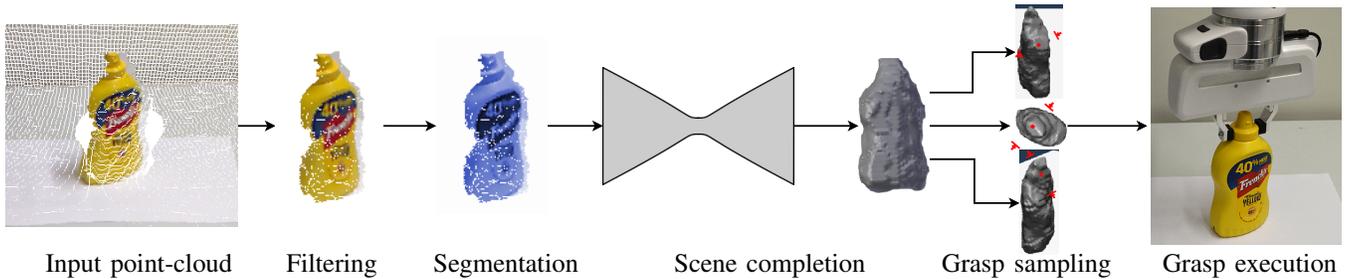
\begin{figure*}
	\centering
	\begin{overpic}[scale=0.46,tics=5]{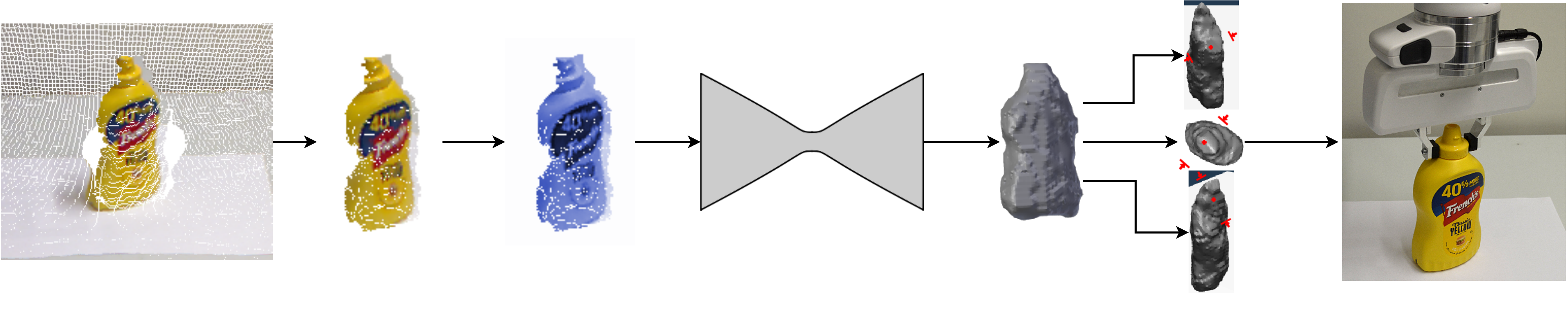}
		\put(3, 0){Input \pc{}}
		\put(21, 0){Filtering}
		\put(32, 0){Segmentation}		
		\put(50, 0){Scene completion}
		\put(70, 0){Grasp sampling}
		\put(86.5, 0){Grasp execution}
	\end{overpic}
	\caption{The proposed grasping pipeline\label{fig:grasping_pipeline}}
	\vspace{-0.7em}
\end{figure*}

In this work, we address the problem of grasping unknown objects lying on a supporting surface with a robotic arm equipped with a parallel-jaw gripper. Information about the scene is obtained by a RGB-D camera whose pose is arbitrary but known relative to the robot.

Formally, let $\matr{x} = (\matr{c}, \matr{O})$ denote a state representing the environment, where $\matr{c} \in \mathbb{R}^6$ is the camera pose, and the set $\matr{O} = \{(\mathcal{O}_i, \matr{p}_i)\}_{i=0}^N$ contains the properties of the $N$ objects to be grasped, described by their pose $\matr{p} \in \mathbb{R}^6$ and model $\mathcal{O}$. The state is partially observable as $\matr{O}$ can only be indirectly and incompletely observed using the RGB-D camera.
The camera produces a 2.5D \pc{} $\matr{y} = \left\{y_i\right\}_{i=1}^{H \times W} \in \mathbb{R}^{H \times W}_+$ which can be represented as a $H \times W$ \depthimage{}, assuming known camera intrinsic parameters.

Let $\matr{g} \in \mathbb{R}^6$ denote a parallel-jaw grasp, described by the 6D pose of the gripper center point and $S(\matr{g}, \matr{x}) \in \{0,1\}$ be a binary-valued grasp success metric indicating, \eg{} force closure. Assuming a joint distribution $p(S, \matr{g}, \matr{x}, \matr{y})$, let $Q(\matr{g}, \matr{y}) = \E{S \mid \matr{g}, \matr{y}}$ be the expected value of the metric given a grasp $\matr{g}$ and a \pc{} $\matr{y}$.
The quality $Q$ is intractable in most real cases. Therefore, it is modeled in data-driven grasp planning methods with a learned parametric model $Q_{\theta}$ with parameters $\theta$. The parametric model $Q_{\theta}$ is typically optimized either with supervised learning on synthetic \cite{mahler2017dex} or real grasping data \cite{varley2015generating}, or with \ac{rl} \cite{levine_learning_2016}.

Then, given a \pc{} $\matr{y}$ obtained from a known camera pose $\matr{c}$, the goal of most data-driven grasp planning methods is to find a grasp $\matr{g}^*$ such that:
\begin{align}\label{eq:gstar}
	\matr{g}^* = \argmax_{\matr{g}\in \mathcal{G}}Q_{\theta}(\matr{g}, \matr{y}),
\end{align}
where $\mathcal{G}$ is a set of grasp candidates. However, such a formulation only accounts for grasps approaching the scene from the same direction as the camera, which is usually looking at it from top, leading to only top-grasps being proposed.

Instead, we propose to extend this framework to allow object grasping from any direction, even those not directly seen by the camera.
For this, we need a function $\hat{\matr{x}} = C(\matr{y})$ as an estimate of the full state $\matr{x}$ from the \pc{} $\matr{y}$. Practically, this means understanding how many and what objects are in the scene (\ie{} segmentation), and then for each object proposing a model $\mathcal{O}$ and a pose $\matr{p}$ (\ie{} shape completion).
Once a state estimate $\hat{\matr{x}}$ is available we want to evaluate the quality of grasps approaching from any direction and execute the first kinetically feasible one with highest quality. Our proposed pipeline to achieve this is described next.

\section{Grasping Pipeline}
\label{sec:method}

The grasping pipeline shown in \figref{fig:grasping_pipeline} consist of:
\begin{enumerate*}[label=(\roman*)]
	\item filtering and segmenting the real objects, 
	\item shape completing each segment and add them to a physics simulator,
	\item generate and rank grasps from different viewpoints of the objects,
	\item execute the best ranked grasp on the real robot.
\end{enumerate*}

\subsection{Segmentation}
\label{sec:segmentation}

The scene in the \pc{} contains an unknown number of objects lying on a table.
We first remove points that are part of the background indicated by their distance exceeding a threshold, as well as points that belong to the supporting surface identified using the known pose of the camera with respect to the robot.
We are then left with a filtered \pc{} $\bar{\matr{y}}$ containing only the view of the objects to be segmented.

Given the \pc{} $\bar{\matr{y}}$, we define an $N$-region segmentation as a partition $\matr{R} =\left\{\matr{r}_i\right\}_{i=1}^N$ of the points of $\bar{\matr{y}}$. More precisely, the regions must satisfy the following constraints:
\begin{align} \label{eq:partition}
   \begin{array}{rl}
      \forall y \in \bar{\matr{y}} & \left( \exists \matr{r} \in \matr{R} \mid y \in \matr{r} \right);\\
      \forall \matr{r} \in \matr{R} & \left( \nexists \matr{r}' \in \matr{R} \setminus \left\{ \matr{r} \right\} \mid \matr{r} \cap \matr{r}' \neq \emptyset \right).
   \end{array}
\end{align}
These constrains enforce that all points in the \pc{} $\bar{\matr{y}}$ have to belong in a region but no point can belong in two regions.

The pipeline is agnostic to the segmentation algorithm employed, but the assumption is that, after the segmentation step, each region in $\matr{R}$ contains points belonging to a different object to be grasped. The next step is to estimate each object's properties through scene completion.

\subsection{Scene Completion}
\label{sec:shape_completion}
Scene completion refers to the process of both shape completing each object in the scene and then placing them in a physics simulator according to their individual estimated pose.
Shape completion refers to reconstructing the shape of an object from partial information about it in the form of a \pc. More precisely, a shape completion algorithm estimates $(\mathcal{O}, \matr{p})$ given a \pc{} $\matr{r}$. To shape complete objects we used the pre-trained fully convolutional hour-glass shaped \ac{dnn} proposed in \cite{lundell2019robust} whose input is a voxel grid of the \pc{} captured from the camera and output is a completed voxel grid. The completed voxel grid is post-processed into a mesh by merging it with the input \pc{} and running the marching cube algorithm \cite{lorensen1987marching}.

The \ac{dnn} in \cite{lundell2019robust} also included dropout layers throughout that were active during run-time to generate a set of shape samples representing, through Monte Carlo sampling, the shape uncertainty. In this work, we also generate shape samples but average them together to get a mean shape, effectively ignoring the shape uncertainty. Although it would be possible to deactivate the dropout layers at run-time and only consider a point estimate of the shape, the benefit of using the mean shape is that it is smoother, removing sharp artefacts on the shape which many \etedd{} methods often rank as stable grasp points. 

For each region $\matr{r}_i \in \matr{R}$ we generate, through shape completion, objects $(\mathcal{O}_i, \matr{p}_i)$. Together, all objects represents an estimate $\hat{\matr{x}}$ of the real environment state $\matr{x}$. The state estimate $\hat{\matr{x}}$, containing all objects represented as meshes, are subsequently placed in a physics simulator. The next step is then to sample grasps over the state estimate.

\begin{figure}
	\centering
	\includegraphics[width=.6\linewidth]{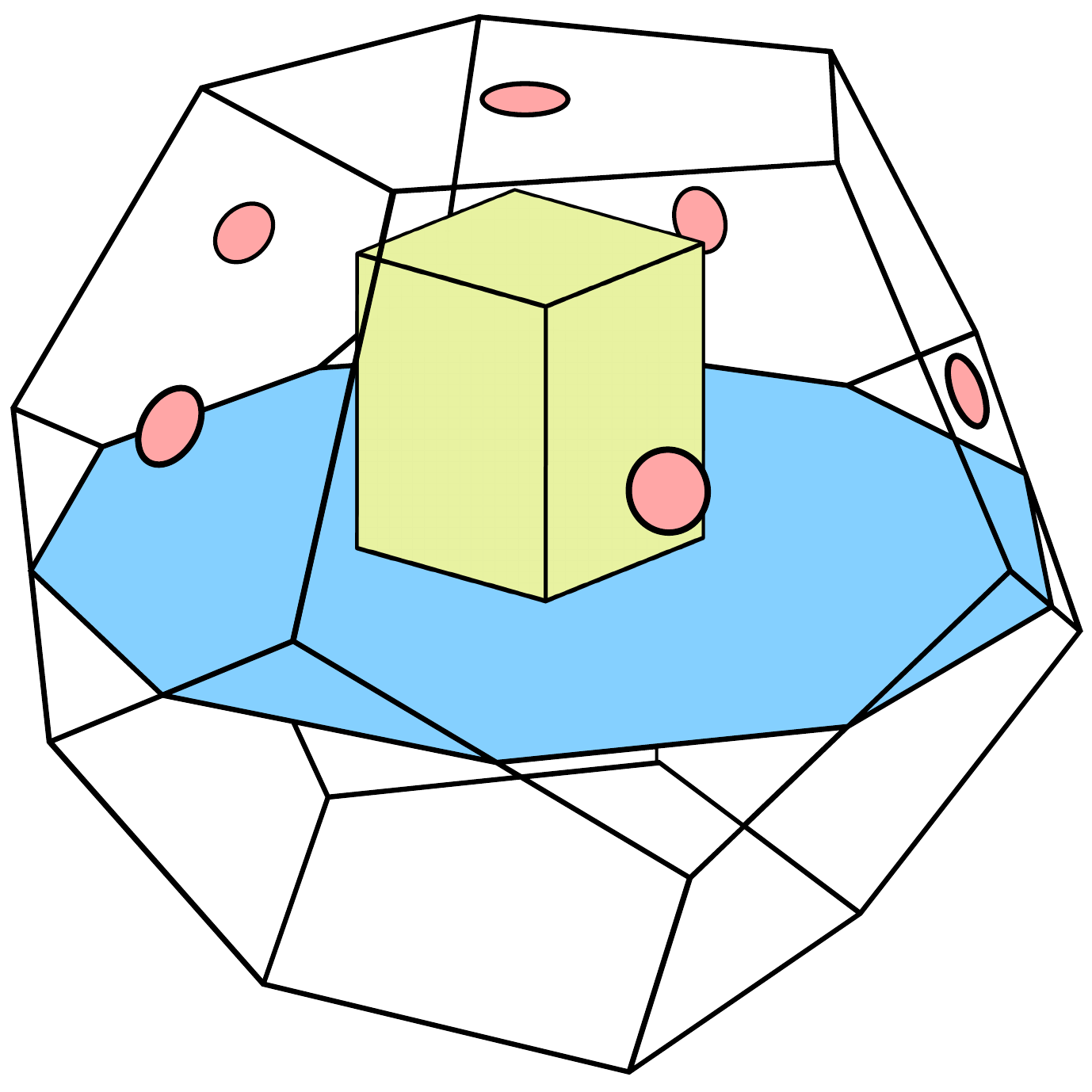}
	\caption{\label{fig:viewpoints}The proposed dodecahedron sampling scheme. The object in yellow is lying on the blue plane. The sampled viewpoints (represented as red circles) are the midpoints of each face in the upper-half of the dodecahedron \bestcolor{}.}
	\vspace{-1em}
\end{figure}

\subsection{Grasp Sampling}
\label{sec:grasp_sampling}

To obtain grasp candidates from all directions, we populate a scene in a physic simulator according to the state estimate $\hat{\matr{x}}$. Given the populated scene, we render $n$ \depthimage{}s $\matr{Y}=\matr{y}_1,~\matr{y}_2,\dots,\matr{y}_n$ of the objects from different viewpoints. To densely sample the scene we propose the sampling scheme visualized in \figref{fig:viewpoints}, which is to approximate a sphere around the workspace with a dodecahedron and use the midpoints of each face in the upper half as the viewpoints. This amounts to $n=6$ viewpoints in total and includes the top-view of the object that most \etedd{} methods are trained on. Of course, other sampling strategies can be devised to obtain an higher or lower number of viewpoints, as desired.

Next, we add noise to each simulated \depthimage{} $\matr{y}_i$ to make them more similar to ones acquired from physical cameras. The reason for adding noise is because many \etedd{} methods \cite{johns_deep_2016,mahler2017dex,satish2019policy} use synthetic data to train $Q_{\theta}$ and adding artificial noise mimics \depthimage{}s acquired from physical cameras which, in turn, improves the sim-to-real transfer. Similar to \cite{mahler2017dex}, we added both multiplicative and additive noise to each viewpoint resulting in the noisy \depthimage{} $\hat{\matr{y}}_i=\alpha\matr{y}_i+\epsilon$, where $\alpha \sim \Gamma(k,s)$ is a Gamma random variable modeling depth-proportional noise, and $\epsilon$ is a pixel-wise zero-mean Gaussian noise as explained in \cite{mahler2015gp} with bandwidth $l$ and measurement variance $\sigma$ modeling additive noise. Experiments verified that adding noise to the depth images made the grasps more robust. 

Grasps $\mathcal{G}_i$ are then generated on the set of noisy \depthimage{}s $\hat{\matr{y}}_i$. The grasp $\matr{g}^*$ that achieves the highest utility among all candidates from all viewpoints is considered the best and, if physically reachable, is executed on the real robot. Formally, the best grasp is
\begin{align}\label{eq:bestg}
	\matr{g}^* = \argmax_{\matr{g}\in \mathcal{G}_i, i = 1,\dots,n} Q_{\theta}(\matr{g}, \hat{\matr{y}}_i).
\end{align}

\section{Experiments}
\label{sec:exp}
The two main questions we wanted to answer in the experiments were:
\begin{enumerate}
	\item What is the impact of generating grasps from simulated \depthimage{}s as opposed to real ones on grasp success and object clearance rate?
	\item Is it beneficial to simulate angled viewpoints instead of top-views only?
\end{enumerate}

In order to provide justified answers to these questions, we conducted two separate experiments. The first experiment evaluates grasp success rate on single object grasping while the second one evaluates the clearance rate in cluttered scenes.

\subsection{Experimental Setup}
\begin{figure}
	\centering
	\begin{overpic}[width=0.45\textwidth,tics=10]{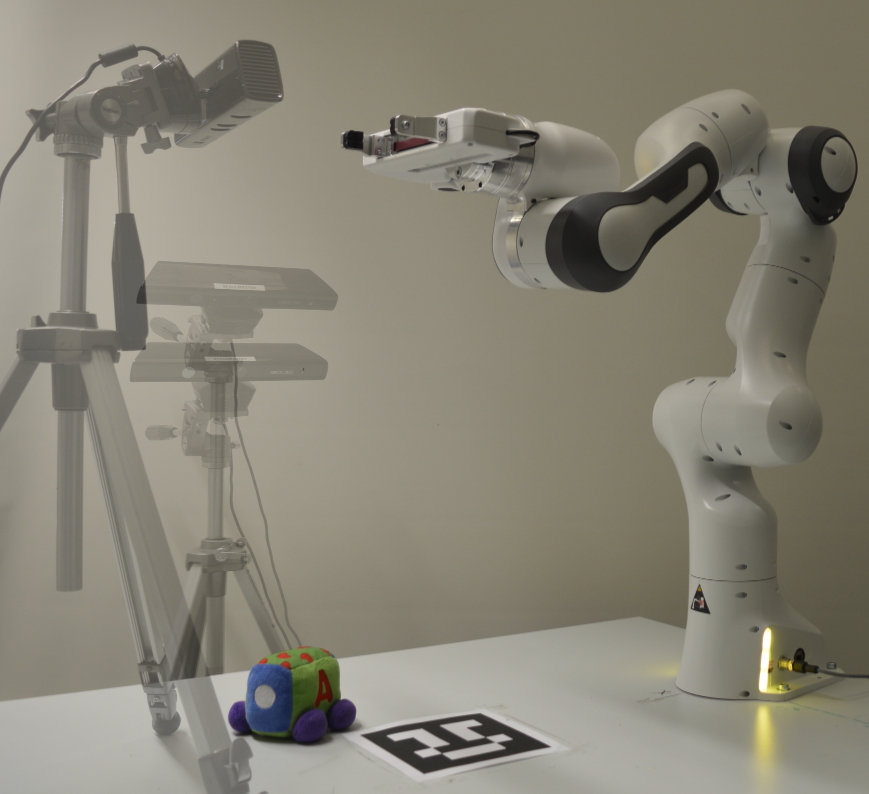} 
		\put(35, 80){90\textdegree}
		\put(40, 56){45\textdegree}
		\put(40, 48){30\textdegree}
	\end{overpic}
	\caption{The three camera viewpoints for single object grasping.\label{fig:setup}}
	\vspace{-1em}
\end{figure}

To perform the experiments we used the \panda{} robot and a Kinect 360\textdegree{} camera to capture the input \pc{}s as shown in \figref{fig:setup}. We used an Aruco marker \cite{garrido2014automatic} for the extrinsic calibration of the camera. Once a \pc{} was captured, it was segmented, shape completed and finally placed into a physical rendering of the scene with the same transformation as in the real world. For segmentation we used the region growing method in PCL and for physical rendering \mujoco \cite{todorov2012mujoco}. For the zero-mean Gaussian noise $\epsilon$ we set $\sigma=0.001$ and kernel bandwidth $l=6$. For the depth-proportional noise $\alpha$, modeled as a Gamma distribution, we set $k=5000$, $s=0.0002$.

In both experiments we tested three different methods all using a pre-trained \ac{fcgqcnn} which is trained to recognize stable top-grasps from depth images \cite{satish2019policy}. The first method, which is the baseline, generates grasps with the \ac{fcgqcnn} on real depth images captured from a Kinect 360\textdegree{} camera. We benchmarked this against our method with two different sampling schemes for simulating depth images: one used the complete dodecahedron sampling method described in \secref{sec:grasp_sampling} while the other sampled a \depthimage{} from a top-down view only. Henceforth we refer to the baseline method as \ac{fcgqcnn}, ours with the dodecahedron sampling as \ac{oag}, and our with top-down sampling as \ac{otg}.

\begin{figure}
	\centering
	\begin{overpic}[width=0.45\textwidth,tics=10]{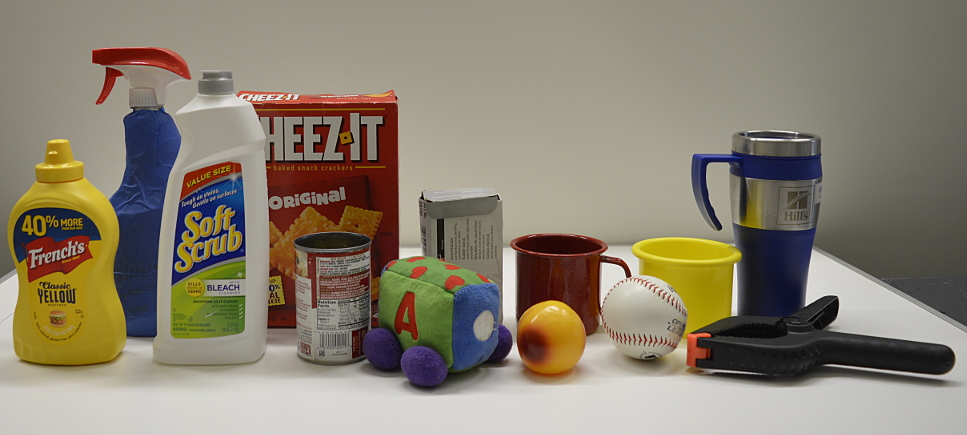} 
		\put(6.5, 1.8){1}
		\put(21.5, 1.8){3}
		\put(33.5, 1.8){4}
		\put(46, 1.8){7}
		\put(56.5, 1.8){8}
		\put(66, 1.8){9}
		\put(80, 1.8){12}
		\put(14, 41){2}
		\put(32, 36){5}
		\put(47, 26){6}
		\put(55.5, 21){10}
		\put(68, 21){11}
		\put(78, 32){13}		
	\end{overpic}
	\caption{The 13 individually numbered objects used in the experiment. All objects, except 6 and 13, are from the YCB object set.\label{fig:objects}}
	\vspace{-1em}
\end{figure}
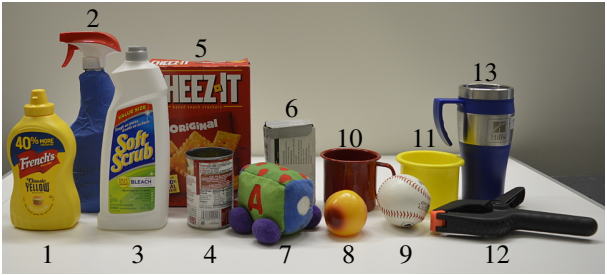

\subsection{Single Object Grasping}
\label{sec:single_object_grasping}

For single object grasping we compared \ac{fcgqcnn} with and without simulated \depthimage{}s on the 13 objects shown in \figref{fig:objects}. To study the effect of the camera viewing angle on grasp performance, we ran the experiments on three different angles towards the grasping plane (30\textdegree{}, 45\textdegree{} and 90\textdegree{}) all shown in \figref{fig:setup}. For the 30\textdegree{} and 45\textdegree{} viewing angles the objects were placed in five different orientations: 0\textdegree{}, 72\textdegree{}, 144\textdegree, 216\textdegree{} and 288\textdegree{}, while for the 90\textdegree{} viewing angle, which corresponds to a top-down view, we only placed the objects at a 0\textdegree{} orientation. In total this setup amounts to 143 grasps per method.

To evaluate if a grasp was successful, the robot moved to the planned grasp pose, closed its fingers, and moved the arm upward 20~cm. Then, the arm moved back to the starting position, and once there rotated the hand $\pm$90\textdegree{} around the last joint. A grasp was successful if the object was within the gripper for this whole procedure and unsuccessful if dropped.

\begin{table*}
	\centering
	\ra{1.4}
	\caption{Average grasp success rate on different viewing angles with test statistics and p-values of pair-wise one sided	Wilcoxon signed-rank test between methods.\label{tb:single_grasping_results}}
	\begin{tabular}{@{}ccccccc@{}}
		\toprule
		Viewing angle& \ac{fcgqcnn} & \ac{otg} & \ac{oag} & \ac{fcgqcnn} vs \ac{otg} & \ac{fcgqcnn} vs \ac{oag}  & \ac{oag} vs \ac{otg}\\
		\midrule
		~~~~~~~~~~~~~~90\textdegree{} (top-down)   & \textbf{69.23\%} & \textbf{69.23\%} & 61.54\%&
		\nat{} & \nat{} & \nat{} \\
		45\textdegree{} & 40.00\% & \textbf{66.15\%} & 58.46\%  &
		\multicolumn{1}{l}{T=52, 
			p\textless{}.0005***} & \multicolumn{1}{l}{T=144, 
			p\textless{}.05*} & \nat{} \\
		30\textdegree{} & 49.23\% & \textbf{67.69\%} & 55.38\% &
		\multicolumn{1}{l}{T=144, 
			p\textless{}.05*} & \nat{} & 
		\nat{} \\ 
				\midrule
		Average success rate &     46.85\%  &  \textbf{67.13\%}  & 57.34\%& \multicolumn{1}{l}{T=450, 
			p\textless{}.0001***} &\multicolumn{1}{l}{T=768, 
			p\textless{}.05*} & \nat{}\\
		\midrule
		Planning  Time  (s) & 2.3 & 2.0 & 12.3 & & &\\
		Shape Completion Time  (s) & \nat{} & 27.6
		 & 27.6
		  & & &\\
		\bottomrule
	\end{tabular}
	\vspace{-0.7em}
\end{table*}

The experimental results for the different methods, which are analyzed for statistical difference with a one sided Wilcoxon signed-rank test, are presented in \tabref{tb:single_grasping_results}. Over all viewing angles, the average grasp success rate is higher with the proposed \ac{otg} and \ac{oag} compared to the baseline \ac{fcgqcnn} ($p<0.0001$, $p<.05$). This result stems from the fact that the performance of \ac{fcgqcnn} deteriorates heavily when moving from a top-down view to an angled view. For instance, the relative performance drop for \ac{fcgqcnn} from a 90\textdegree{} viewing angle to a 45\textdegree{} is -42.22\% and to 30\textdegree{} the drop is -28.89\%. This is much higher compared to the performance drop for \ac{oag}, which is only -5\% and -10\%. The performance drop for \ac{otg} is even less with -4.4\% and -2.2\% when moving from a 90\textdegree{} viewing angle to a 45\textdegree{} and 30\textdegree{} respectively. Together, these results show the importance of simulating \depthimage{}s if the viewing angle of the real camera is not 90\textdegree{}.

Another interesting result from \tabref{tb:single_grasping_results} is that \ac{otg}, which simulates only top-down views, outperforms \ac{oag} which, in addition to simulating a top-down view, also simulates from angled viewpoints. One reason \ac{otg} achieved a higher grasp success rate than \ac{oag} was that in many cases when an angled grasp was executed the gripper either tilted the object over or if the gripper decided to grasp a corner of an object the object simply slipped out of the gripper. Such situations were not common for top-grasps as the surface on which the object lies prevents the object from slipping and reduces the chance of it falling over. Although top-grasps seem more robust to external perturbations, we hypothesize that the performance difference between \ac{otg} and \ac{oag} could be reduced if \ac{fcgqcnn} was also trained on angled viewpoints.

\begin{figure*}
	\centering
	\begin{tikzpicture}
\begin{axis}[
	ymin=-26, ymax=45,
	ybar=0pt,
	bar width=9pt,
	x=1.26cm,
	xmin=1, xmax=13,
	enlarge x limits={.04},
	ylabel={Grasp success rate -- average sucess rate per object},
	xlabel={Objects},
	axis x line*=bottom,
	axis y line*=left,
	ymajorgrids,
	tick pos=left,
	y grid style={white!50.0!black},
	legend style={
		at={(0.5,1.1)},
		anchor=north,
		legend columns=-1,
	},
	symbolic x coords={1,2,3,4,5,6,7,8,9,10,11,12,13},
	xtick=data,
	nodes near coords,
	nodes near coords style={rotate=90, anchor=west}, 
	point meta=explicit symbolic
]

\addplot+ coordinates{
	(1,-8.33) 
	(2,-25) 
	(3,-22.22) 
	(4,-2.78) 
	(5,-8.33) 
	(6,16.67) [16.67]
	(7,2.78) [2.78]
	(8,-19.44) 
	(9,0) [0]
	(10,-11.11) 
	(11,-11.11) 
	(12,-25) 
	(13,-8.33) 	
};
\begin{scope}[every node/.style={rotate=90, anchor=west, color=blue, yshift=9pt}]
	\node at (1,0) {$-8.33$};
	\node at (2,0) {$-25$};
	\node at (3,0) {$-22.22$};
	\node at (4,0) {$-2.78$};
	\node at (5,0) {$-8.33$};
	\node at (8,0) {$-19.44$};
	\node at (10,0) {$-11.11$};
	\node at (11,0) {$-11.11$};
	\node at (12,0) {$-25$};
	\node at (13,0) {$-8.33$};
\end{scope}

\addplot+ coordinates{
	(1,33.34) [33.34]
	(2,16.67) [16.67]
	(3,-5.56) 
	(4,5.55) [5.55]
	(5,16.67) [16.67]
	(6,0) [0]
	(7,2.78) [2.78]
	(8,13.89) [13.89]
	(9,8.33) [8.33]
	(10,5.55) [5.55]
	(11,5.55) [5.55]
	(12,16.66) [16.66]
	(13,0) [0]
};
\begin{scope}[every node/.style={rotate=90, anchor=west, color=red}]
	\node at (3,0) {$-5.56$};
\end{scope}

\addplot+ coordinates{
	(1,-25) 
	(2,8.34) [8.34]
	(3,27.78) [27.78]
	(4,-2.78) 
	(5,-8.33) 
	(6,-16.67) 
	(7,-5.55) 
	(8,5.56) [5.56]
	(9,-8.34) 
	(10,5.55) [5.55]
	(11,5.55) [5.55]
	(12,8.33) [8.33]
	(13,8.33) [8.33]
};
\begin{scope}[every node/.style={rotate=90, anchor=west, color=brown!60!black, yshift=-9pt}]
	\node at (1,0) {$-25$};
	\node at (4,0) {$-2.78$};
	\node at (5,0) {$-8.33$};
	\node at (6,0) {$-16.67$};
	\node at (7,0) {$-5.55$};
	\node at (9,0) {$-8.34$};
\end{scope}

\legend{\ac{fcgqcnn},\ac{otg},\ac{oag}}
\end{axis}
\end{tikzpicture}
	\vspace{-1em}
	\caption{Grasp success rate per object for each method minus the average success rate per object on each of the 13 objects used in the experiment.\label{fig:individual_obj}}
	\vspace{-1em}
\end{figure*}
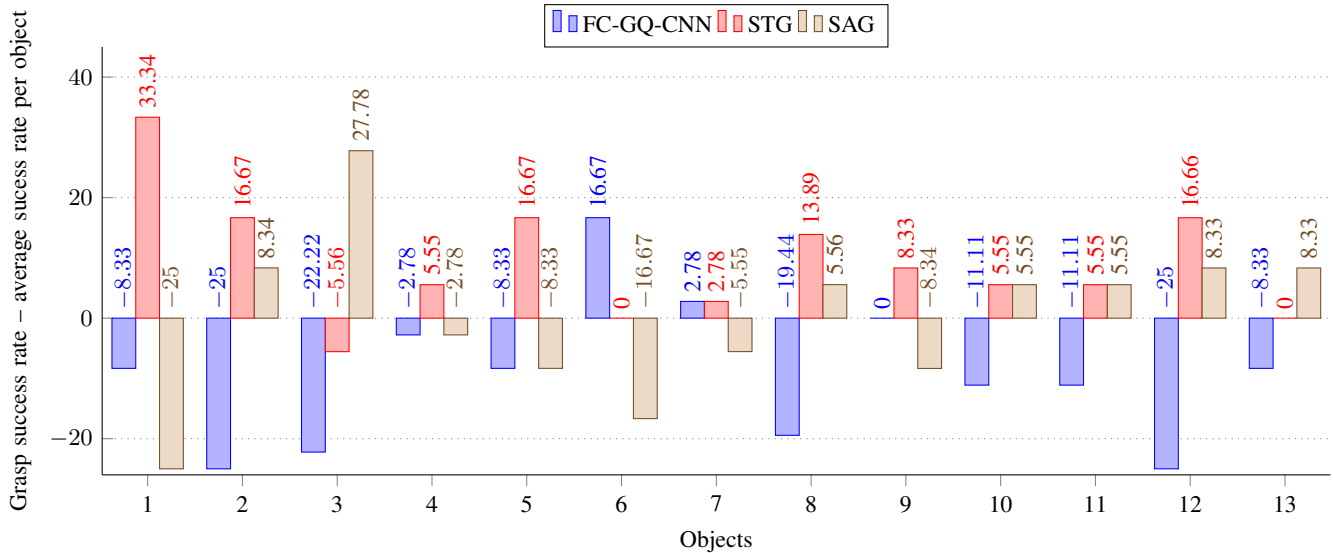

Finally, \figref{fig:individual_obj} shows clearly that \ac{otg} performs better than average on all objects except for object 3 while \ac{oag} is above average on 7 out of the 13 objects. The performance of \ac{fcgqcnn}, on the other hand, is worse than average on 10 objects with an over 20\% worse than average performance on objects 2, 3, and 12. One possible reason \ac{fcgqcnn} performs poorly on those objects is that grasping them from an angled viewpoint is much harder than grasping them from the top.

\subsection{Grasping in Clutter}
\label{sec:grasping_in_clutter}

In this experiment we studied the clearance rate of each method in a cluttered scene, meaning that the objective was to remove as many objects as possible within a given grasping budget. The grasping budget was set to 12 grasps and the objects we chose to use were 4, 6, 7, 8, 10 and 12 in \figref{fig:objects} as these represent different shapes and sizes. To generate a cluttered scene, the objects were placed in a box that was shaken and emptied onto a table. An example scene is shown in \figref{fig:cluttered_scene}. The physical camera perceiving the scene was set to 45\textdegree{}. To evaluate if an object was successfully removed from the scene we used the same procedure as in the single object grasping experiments except the last step to rotate the gripper was excluded for speed.

The experimental results are presented in \tabref{tb:grasping_in_clutter}. These results show a clear improvement in average clearance rate using \ac{otg} and \ac{oag} over \ac{fcgqcnn}. For instance, \ac{otg} removed all objects in 9 out of 10 scenes and for the one scene it did not clear only one object was left. \ac{oag}, on the other hand, managed to clear 5 out of 10 scenes while \ac{fcgqcnn} cleared 2 out of 10 scenes. 

For scenes where not all objects were cleared, the average clearance rate were 83.33\% for \ac{otg}, 76.68\% for \ac{oag}, and 47.9\% for \ac{fcgqcnn}. In these cases, \ac{fcgqcnn} manged to remove more than half of the objects in only 2 out of the 8 scenes. \ac{oag} and \ac{otg}, on the other hand, removed more than half of the objects in all scenes.

Based on the presented results, we demonstrated that it is also beneficial to generate grasps from other viewing angles when removing objects in cluttered scenes. Together, both the result on single object grasping and grasping in clutter demonstrate that the performance of \ac{fcgqcnn}, a \sota{} \etefdd{} method, deteriorates heavily when viewing the scene from an angled viewpoint. However, through the use of shape completion and simulated viewpoints this is no longer the case.

\begin{table}
	\centering
	\ra{1.3}\tbs{10}
	\caption{\label{tb:grasping_in_clutter}Results on the cluttered scene}
	\begin{tabular}{@{}lccc@{}}
		\toprule
		& \ac{fcgqcnn} & \ac{otg} & \ac{oag} \\
		\midrule
		Average clearance rate (\%) & 58.33 & \textbf{98.33} & 88.33\\
		Planning  Time (s) & \textbf{1.86} & 2.7 & 12.84 \\
		Shape Completion Time (s) & \nat{} & 47.53   & 60.23\\
		\bottomrule                                                         
	\end{tabular}
	\vspace{-1em}
\end{table}

\section{Conclusions}
\label{sec:concl}
We presented a grasping pipeline that enables \etedd{} methods which previously only generated 4 \acp{dof} top-grasps from a single \depthimage{} to generate full 6 \acp{dof} grasps from simulated viewpoints. The key component was the use of shape completion to model a partly observed object and place it into a physics simulator to simulate \depthimage{}s from multiple viewpoints. We used \ac{fcgqcnn} to generate grasps and compared the 6 \acp{dof} grasps generated with our pipeline to the 4 \acp{dof} grasps proposed from a \depthimage{} captured by a real camera on both single object grasping and grasping in clutter. The single object grasping results show that generating full 6 \acp{dof} grasps leads to a statistical significant improvement in terms of higher grasp success rate. Major improvements were also prominent for grasping in clutter when generating 6 \acp{dof} grasps opposed to 4 \acp{dof} ones.

Despite the good results, shape completion is a major computational bottleneck. Most computation time, however, is not spent on shape completion but on the post-processing of the completed voxel grid which could be improved with better hardware and optimized implementation. Another limitation is that the accuracy of shape completion is conditional on successful segmentation. The analytical region growing segmentation method used here is known to perform poorly in highly cluttered scenes \cite{danielczuk2019segmenting}. Thus, the segmentation method would need to be replaced in such cases.

In conclusion the work presented here demonstrates that planning full 6 \acp{dof} grasps brings significant advantages over 4 \acp{dof} grasps. This, in turn, poses new interesting research questions. For instance, instead of simulating different viewpoints of the shape completed object, is it maybe better to plan directly on the object itself using, \eg{} mesh neural networks \cite{feng2019meshnet}? Or, is it possible to generate full 6 \acp{dof} grasps directly from real \depthimage{}, removing the need for shape completion? These  questions pave way for interesting future research avenues.


\section*{Acknowledgment}

We gratefully acknowledge the support of NVIDIA 
Corporation with the donation of the Titan Xp GPU used for this research.



\bibliographystyle{IEEEtran}
\bibliography{refs}

\end{document}